\documentclass[letterpaper, 10pt, conference]{ieeeconf}
\IEEEoverridecommandlockouts                              
\usepackage{graphics} 
\usepackage{epsfig} 
\usepackage{mathptmx} 
\usepackage{times} 
\usepackage{amsmath} 
\usepackage{amssymb}  
\usepackage{algorithmic}
\usepackage[ruled, vlined, linesnumbered]{algorithm2e}
\usepackage{comment}
\usepackage{balance} 
\usepackage{subcaption}

\title{\LARGE \bf
Reward Shaping with Subgoals for Social Navigation
}

\author{Takato Okudo$^{1}$ and Seiji Yamada$^{2}$
\thanks{$^{1}$Takato Okudo is with The Graduate University for Advance Studies, SOKENDAI, Tokyo, Japan {\tt\small okudo@nii.ac.jp}}%
\thanks{$^{2}$Seiji Yamada is with National Institute of Informatics, NII, Tokyo, Japan {\tt\small seiji@nii.ac.jp}}%
}

\begin{document}

\maketitle
\thispagestyle{empty}
\pagestyle{empty}

\begin{abstract}
	Social navigation has been gaining attentions with the growth in machine intelligence. Since reinforcement learning can select an action in the prediction phase at a low computational cost, it has been formulated in a social navigation tasks. However, reinforcement learning takes an enormous number of iterations until acquiring a behavior policy in the learning phase. This negatively affects the learning of robot behaviors in the real world. In particular, social navigation includes humans who are unpredictable moving obstacles in an environment. We proposed a reward shaping method with subgoals to accelerate learning. The main part is an aggregation method that use subgoals to shape a reinforcement learning algorithm. We performed a learning experiment with a social navigation task in which a robot avoided collisions and then reached its goal.
	The experimental results show that our method improved the learning efficiency from a base algorithm in the task.
\end{abstract}

\section{INTRODUCTION}
A society in which social robots actively work with humans is approaching with the growth of machine intelligence. For this, the design of robots is required to take human appearances into consideration. In addition, since the navigation tasks are basic and applicable to areas involving humans, social navigation has attracted a lot of researchers~\cite{Kruse:2013:HRN:2542686.2542722, Che2020Efficient, Heiden2020Social}. According to Kruse et al.~\cite{Kruse:2013:HRN:2542686.2542722}, the studies in the field of social navigation have tried to make interaction more comfortable for humans, and there are broadly three goals: comfort, naturalness, and sociability. For the first goal, the robot behaves in such a way so as to make humans feel comfortable so that an interaction can proceed without annoyance and stress. For the second, the robot has a naturalness in its low-level behavior patterns that is similar to humans. For the third, robots with sociability adhere to explicit high-level cultural conventions. There are several well-engineered methods for satisfying all goals~\cite{Helbing1995Social, 10.1007/978-3-642-19457-3_1}. 
They must rely on manual modeling, which is laborious. In contrast to such approaches, modeling in a data-driven fashion has been focused on~\cite{Chen2019Crowd, Heiden2020Social}. In particular, deep reinforcement learning(DRL) approaches are useful because a learnt model can generate a human-aware path with little computational costs. 
However, reinforcement learning is time-consuming for model learning. Accelerating learning is important to applying reinforcement learning to acquire robot behaviors. In robotics, in particular, there is often a gap between simulation and the real world, so fine-tuning is required~\cite{Kaspar2020Sim2RealTF}. Incorporating human knowledge into learning algorithms is useful~\cite{ijcai2018-817}. Though there are a lot of studies on using trajectories~\cite{Ng:2000:AIR:645529.657801}, we focus on human knowledge of subgoals. A teacher of trajectory knowledge needs skills in robot control to generate trajectories. A teacher of subgoal knowledge only needs to consider the best trajectory. This might mean less requirements for teachers of intelligent robots. To incorporate subgoal knowledge into RL, we use a reward shaping method to accelerate learning. Since the transformation of rewards must not change the original optimal policy, we make the potential-based reward shaping(PBRS) the basis~\cite{Ng+HR:1999}. A potential of SARSA-RS~\cite{Grzes2008} is defined as a state-value over high-level states. We use SARSA-RS as the basis because it excludes designing the size of potentials. The teacher can focus on what subgoals are in the RL task. 
The main contributions of this work are (i) the dynamic trajectory-based aggregation for SARSA-RS, (ii) improvement to the learning efficiency for CADRL, (iii) a definition for a subgoal of a social navigation task.

\section{Related Work}

Social navigation for robots that can behave and move in a socially-compliant manner in environments where humans have actions, is becoming the most important research direction in human-robot interaction (HRI). Recently, various studies have been done in this field including mobile robots, path planning, and machine learning. 

Che et al.~\cite{Che2020Efficient} proposed a path planning framework that can utilize both implicit expression (robot motion) and explicit expression (visual feedback) so that humans and mobile robots can operate in the same environment without conflict. They developed a model of a human that deals with both continuous movements and discrete behavior in human navigation. The model assumes that the human is a rational agent, with a reward function obtained through inverse reinforcement learning. Then, a planner uses this model to generate communicative actions that maximize the robot's transparency and efficiency. They experimentally found that the planner generated plans that were easier for a human to understand, and the users' trust was increased in comparison with simple collision avoidance.

Chen et al.~\cite{Chen2019Crowd} proposed a deep reinforcement learning method for acquiring effective movement strategies to operate in a socially-compliant way in crowded spaces. The advantage of their work is that the deep reinforcement learning method is based on an attention mechanism that pools attentions and models both of human-human and human-robot interaction. Also, high-ordered crowd-robot interaction was introduced here. Various experiments were conducted, and the results demonstrate that their model anticipated human dynamics and navigated in crowds with time efficiency, outperforming state-of-the-art methods. 

Heiden et al. improved the quality of social navigation by introducing human empowerment and a deep reinforcement learning technique~\cite{Heiden2020Social}. Human empowerment is a kind of bias and set of evaluation functions for learning effective behaviors to avoid getting in the way of a human in workspaces that humans and robots share. They successfully used the empowerment function to make a mobile robot's navigation more social and comfortable to humans. Also, their technique made learning feasible. They conducted simulation-based experiments, where a robot and several humans approached their own independent and dynamic goals. They found that human empowerment was able to improve the performance and sociality of the robot's navigation. 

Kivrak et al. proposed a novel social navigation framework for mobile service robots, maintaining both human safety and comfort in environments where humans perform actions~\cite{Kivrak2020Social}. Their main contribution is that the framework can be used for unknown environments. To achieve their goal, they developed a pedestrian model based on a social force model~\cite{Helbing1995Social}. This model works well, particularly for low-density environments, and is employed as a local planner to generate human-friendly plausible routes for our service robot in corridor-like indoor environment scenarios. They fully implemented the framework as ROS nodes, and they experimentally evaluated the framework both in real world and simulation environments and successfully obtained positive results.

Although several studies have tried to utilize human knowledge including implicit user feedback and optimal paths, few studies have introduced human knowledge, including subgoals to improve reinforcement learning's efficiency. 

There have been various studies that have tried to improve reinforcement learning's efficiency by introducing human knowledge such as subgoals in general tasks. 

The landmark-based reward shaping of Demir et al.~\cite{Demir2019} is the closest to our method. Their method shapes only rewards on a landmark by using a value function. Their study focused on a POMDP environment, and landmarks automatically become abstract states. We focus on an MDP environment, and we propose an aggregation function. We acquire subgoals from human participants, and we apply our method to a task with high-dimensional observations. 
Harutyunyan et al.~\cite{Harutyunyan2015} has shown that the q-values learned by arbitrary rewards can be used for potential-based advice. The method mainly assumes that a teacher negates the agent's action selection. It uses failures in the trial and errors. In contrast, our method uses successes. \par
Reward shaping in HRL has been studied in~\cite{Gao2015, Li2019}. Gao et al.~\cite{Gao2015} showed that potential-based reward shaping remains policy invariant to the MAX-Q algorithm. Designing potentials every level is laborious work. We use a single high-level value function as a potential, which reduces the design load. Li et al.~\cite{Li2019} incorporated an advantage function in high-level state-action space into reward shaping. Their approach is similar to ours in the utilization of a high-level value function, but they do not incorporate external knowledge into their algorithm. The reward shaping method in~\cite{Paul2019} utilized subgoals that are automatically discovered with expert trajectories. The potentials generated every subgoal are different. The value of a potential is fixed and not learned. Our method learns the value of a potential.


\section{Preliminary and Background}
A Markov decision process consists of a set of states $S$, a set of actions $A$, a transition function $T: S\times A \rightarrow (S' \rightarrow [0,1])$, and a reward function $R: S\times A \rightarrow \mathcal{R}$. A policy is a probability distribution over actions conditioned on states, $\pi:S\times A \rightarrow [0, 1]$. In a discounted fashion, the value function of a state~$s$ under a policy $\pi$, denoted $v_\pi(s)$, is $V_\pi = \mathbf{E}_\pi \left[ \sum_{i=0}^\infty \gamma^i r_{t+1} | s_0=s \right]$. Its action-value function is $Q(s,a) = \mathbf{E}_\pi \left[ \sum_{i=0}^\infty \gamma^i r_{t+1} | s_0=s, a_0=a \right]$, where $\gamma$ is a discount factor.
\subsection{Potential-Based Reward Shaping}
Potential-based reward shaping~\cite{Ng+HR:1999,DBLP:journals/corr/abs-1106-5267} is an effective method for keeping an original optimal policy $\pi$ in an environment with an additional reward function $F$. If the potential-based shaping function $F$ is formed as
\begin{eqnarray*}
	F(s_t,s_{t+1}) = \gamma \Phi(s_{t+1}) - \Phi(s_t)
\end{eqnarray*}
, it is guaranteed that policies~$\pi$ in MDP $M=(S,A,T, \gamma, R)$ are consistent with those $\pi'$ in MDP $M'=(S,A,T,\gamma, R+F)$. Note that $s_t \in S-\{s_0\}$ and $s_{t+1} \in S$. $s_0$ is an absorbing state, so the MDP ``stops" after a transition into $s_0$. $\Phi$ is known as the potential function. $\Phi$ should be a real-value function such as $\Phi: S \rightarrow \mathbf{R}$. For better understanding, we use the example of Q-learning with potential-based reward shaping. The learning rule is formally written as
\begin{gather*}
	Q(s_t,a_t) \leftarrow Q(s_t, a_t) +\alpha\delta_{TD} \\
	\delta_{TD} = r_t + F(s_t, s_{t+1}) + \max_{a'} Q(s_{t+1}, a_{t+1}) - Q(s_t,a_t)
\end{gather*}

, where $\alpha$ is a learning rate. We need to define an appropriate $\Phi$ for every domain. There is the problem of how to define $\Phi$ to accelerate learning. The study of~\cite{DBLP:journals/corr/abs-1106-5267} has shown that learning with potential-based reward shaping is equivalent to $Q$-value initialization with the potential function$\Phi$ before learning. The result has made clear that $\Phi(s) = V^*(s) = \max_{a'} Q^*(s, a')$ is the best way to accelerate learning. We cannot know $V^*(s)$ before learning since we acquire $V^*(s)$ after learning. This suggests that we can accelerate learning if there is a value function learned faster than another with the same rewards.

\subsection{SARSA-RS}
Grzes et al.~\cite{Grzes2008} proposed a method that learns a potential function $\Phi$ during the learning of a policy $\pi$, called  ``SARSA-RS." The method solved the problem of the design of an appropriate potential function for a domain being too difficult and time-consuming. We define $Z$ as a set of abstract states. The method builds a value function over $Z$ and uses it as $\Phi$:
\begin{eqnarray*}
	\Phi(s) = V(g(s)) = V(z)
\end{eqnarray*}
, where $g$ is an aggregation function, $g: S \rightarrow Z$. The function $g$ is pre-defined. The potential-based shaping function over SARSA-RS is written as follows.
\begin{eqnarray*}
	F(z_t, z_{t+1}) = \gamma V(z_{t+1}) - V(z_t)
\end{eqnarray*}
The method learns the value function $V(z)$ during policy learning as:
\begin{eqnarray*}
	V(z_t) \leftarrow V(z_t) + \alpha \left( r_h + \gamma^k V(z_{t+1}) - V(z_t) \right)
\end{eqnarray*}
, where $r_h$ is a function for transforming MDP rewards into SMDP rewards, and $k$ is the duration between $z_{t}$ and $z_{t+1}$. The potential function changes dynamically during learning, and the equivalency of the potential-based reward shaping cannot be applied because it depends on the time in addition to the state.
Since Devlin and Kudenko have shown that a shaped policy is equivalent to a non-shaped one, when the potential function changes dynamically during learning~\cite{Devlin:2012:DPR:2343576.2343638}, SARSA-RS keeps the learned policy original. We omit the time argument in the following section to simplify the expression. The size of $Z$ is smaller than $S$ thanks to the aggregation of states. Therefore, the propagation of environmental rewards is faster, and the policy learning with SARSA-RS is also faster. As mentioned above, the method requires the pre-defined aggregation function $g$. In an environment of high-dimensional observations, it is almost impossible to make an aggregation function.
\subsection{Collision Avoidance with Deep Reinforcement Learning}
The multi-agent collision avoidance problem can be formulated in an RL framework~\cite{7989037, 8202312}. A state $s$ consists of observable and hidden parts, that is, $\mathbf{s} = [\mathbf{s}^o, \mathbf{s}^h]$. The observable part $\mathbf{s}^o$ denotes the state that can be measured by all other agents. Let the observable states be the agent's position, velocity, and radius, $\mathbf{s}^o=[p_x, p_y, v_x, v_y, r] \in \mathbb{R}^5$, and let the hidden states be the agent's intended goal position, preferred speed, and heading angle, $\mathbf{s}^h=[p_{gx}, p_{gy}, v_{pref}, \psi] \in \mathbb{R}^4$, and let action $a$ be the agent's velocity. The reward function is as follows.
$$
R(s, a) =
	\begin{cases}
		-0.25& \rm{if} \ d_{min} < 0 \\
		-0.1-\rm{d_{min}}/2 & \rm{else\ if} \ d_{min} < 0.2\\
		1& \rm{else\ if} \ \mathbf{p}=\mathbf{p}_g\\
		0& \rm{otherwise}
	\end{cases}
$$
, where $\rm{d_{min}}$ is the minimum distance of separation between the two agents during $\delta t$, given the agent's velocity $a = \left[ v_s, \phi \right]$, $\mathbf{p}$ is the agent's position on a 2D plane, $\mathbf{p}_g$ is the position of the goal, and $s$ is a combination of the agent's own state~$\mathbf{s}$ and the observable state of the other~$\mathbf{\tilde{s}}^o$, $s = [\mathbf{s}, \mathbf{\tilde{s}}^o]$. The reward function returns a positive evaluation for achieving goals, and a negative one for collisions and shorter distances to the other agent than a threshold. \par
The CADRL algorithm uses a deep neural network as a value function. The network is trained to minimize a quadratic regression error, $\arg\min_{\theta}\sum^{N}_{k=1}\left(y_k - V(s_k; \theta)\right)^2$. The policy selects an action with constraints of
\begin{eqnarray}
	\label{eq:angle}
	\pi(s) = a = [v_s, \phi] \  \rm{for} \ v_s < v_{pref}, |\phi - \psi| < \frac{\pi}{6} \\
	\label{eq:heading}
	|\psi_{t+1} - \psi_t|<\delta t\cdot v_{pref}
\end{eqnarray}
, where (\ref{eq:angle}) limits the direction the agent can rotate in, and (\ref{eq:heading}) specifies the maximum angle at which the agent can rotate. CADRL has an initialization step. In this step, trajectories are generated by ORCA, which is not an optimal policy, but a goal can be reached. The value network is trained with the trajectories. After this step, CADRL starts the learning step. In the experiment, we use the CADRL algorithm as the basis.

\section{Reward Shaping with Subgoal-Based Aggregation}
\label{sec:reward-shaping}
We propose a method of aggregation from states into an abstract states. The method basically follows SARSA-RS. We use a pre-defined subgoal series and aggregate episodes dynamically into abstract states during learning with it. 

\subsection{Subgoal}
We define a {\it subgoal} as a state $s$ if $s$ is a goal in one of the sub-tasks decomposed from a task. In the option framework, the subgoal is the goal of a sub-task, and it is expressed as a termination function~\cite{Sutton:1999:MSF:319103.319108}. Many studies on the option framework have developed automatic subgoal discovery~\cite{Bacon:2017:OA:3298483.3298491}. We aim to incorporate human subgoal knowledge into the reinforcement learning algorithm with less human effort required. The property of a subgoal might be a part of the optimal trajectories because a human should decompose a task to achieve a goal. We acquire a subgoal series and incorporate subgoals into our method in the experiment. The subgoal series is written formally as $(SG, \prec)$. $SG$ is a set of subgoals and a sub-set of $S$. There are two types of subgoal series, totally ordered and partially ordered. With totally ordered subgoals, a subgoal series is deterministically determined at any subgoal. In contrast, partially ordered subgoals have several transitions to the subgoal series from a subgoal. We used only the totally ordered subgoal series in this paper, but both types of ordered subgoals are available for our proposed reward shaping. Since an agent needs to achieve a subgoal only once, the transition between subgoals is unidirectional. 
\subsection{Subgoal-Based Dynamic Trajectory Aggregation}
We propose a method of aggregating trajectories dynamically into abstract states using subgoal series. The method makes the SARSA-RS method available for environments of high-dimensional observations thanks to less effort being required from designers. The method requires only a subgoal series consisting of several states instead of all states. In this section, we assume that the subgoal series $(SG, \prec)$ is pre-defined, and $(SG, \prec) = \left\{sg_0 \prec sg_1 \prec \cdots \prec sg_n \right\}$. The method basically follows SARSA-RS, and the difference is mainly the aggregation function $g$ and minorly the accumulated rewards.
\subsubsection{Dynamic Trajectory Aggregation}
\begin{figure}[tb]
	\centering
	\begin{subfigure}[b]{0.235\textwidth}
		\centering
		\includegraphics[width=\textwidth]{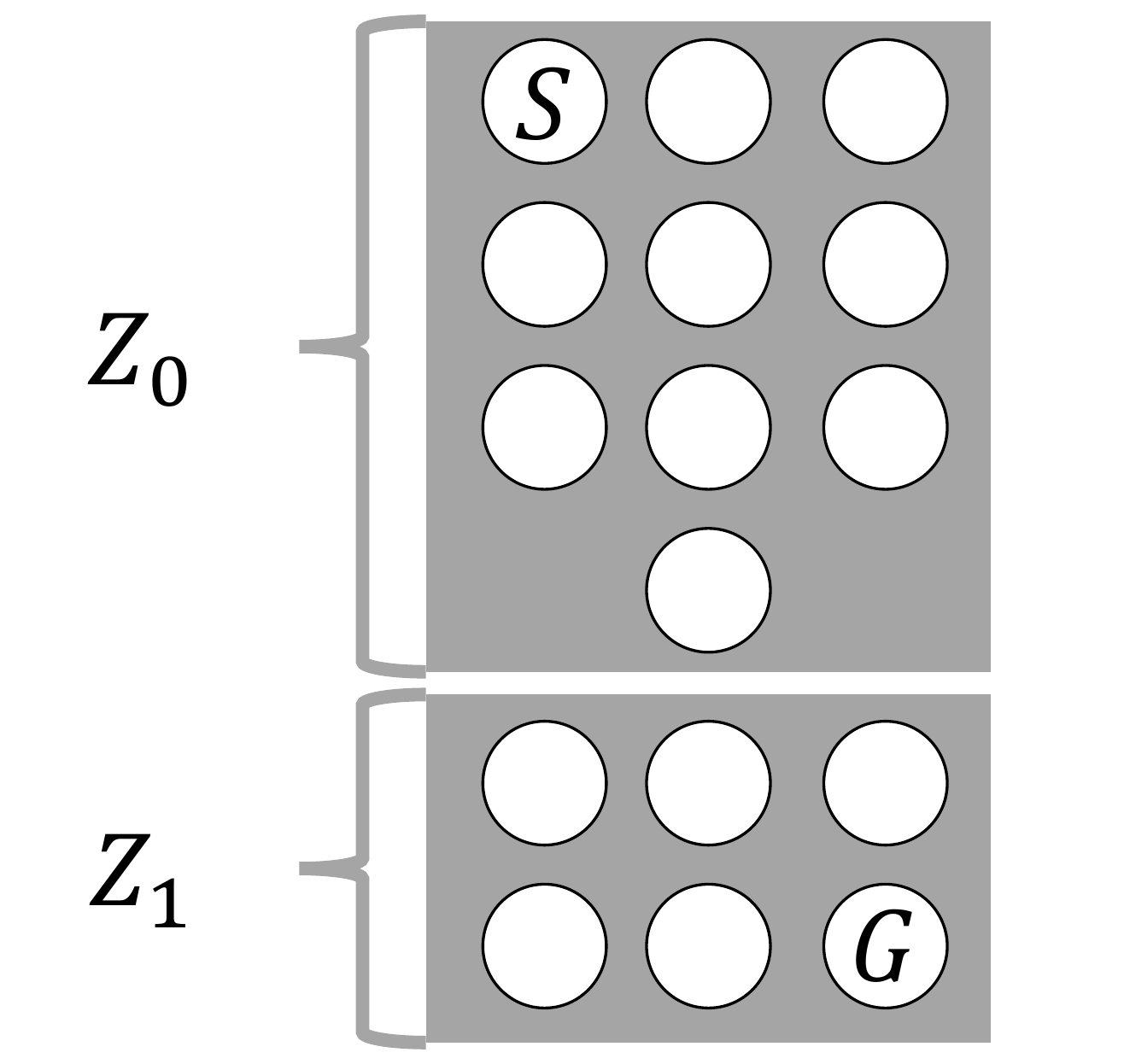}
		\caption{Aggregation function in SARSA-RS}
		\label{fig:sarsa-rs-concept}
	\end{subfigure}
	\hfill
	\begin{subfigure}[b]{0.235\textwidth}
		\centering
		\includegraphics[width=\textwidth]{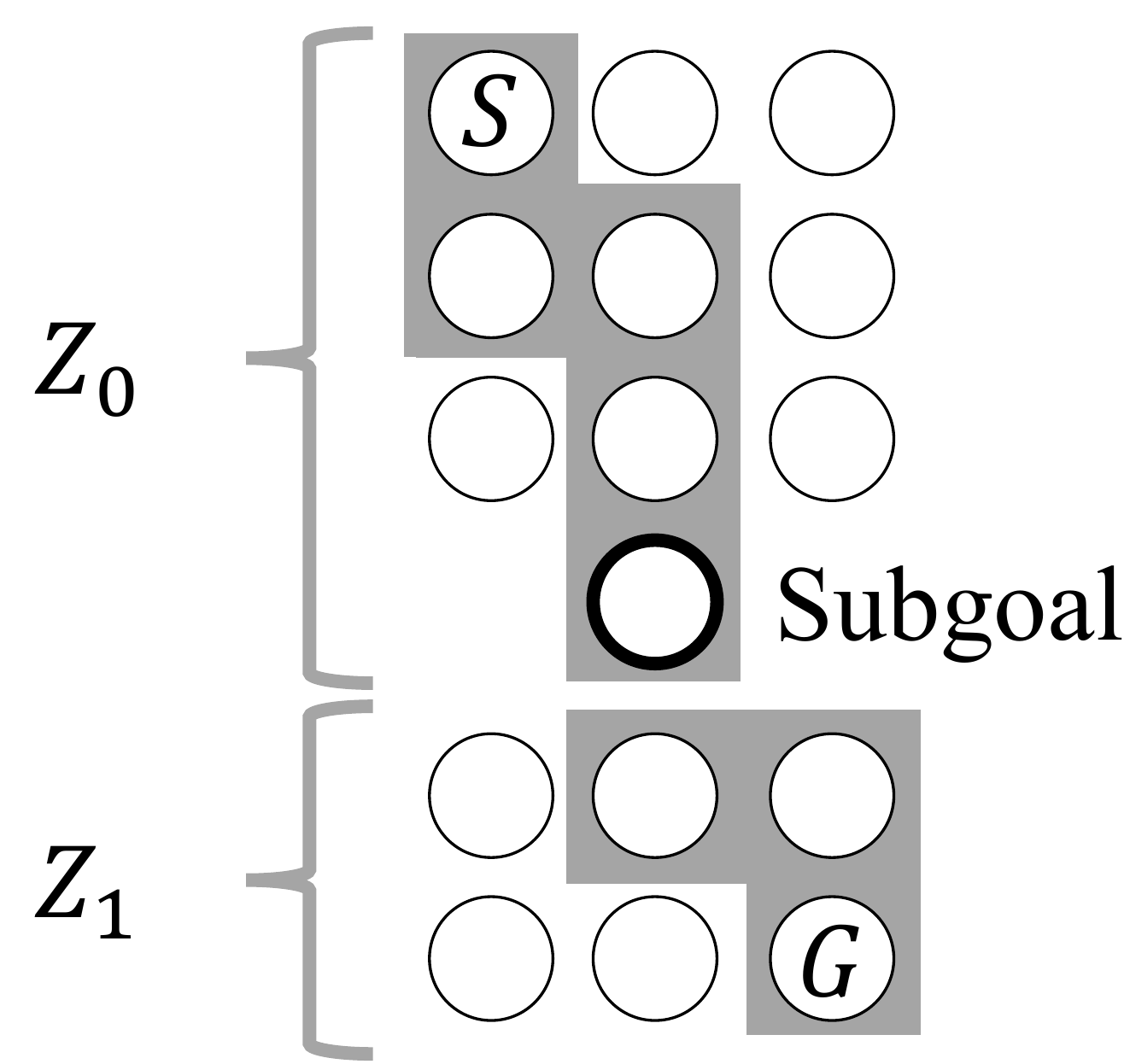}
		\caption{Subgoal-based dynamic trajectory aggregation}
		\label{fig:dta}
	\end{subfigure}
	\caption{Concept of subgoal-based aggregation}
	\label{fig:concept}
\end{figure}

We build abstract states to represent the achievement status of a subgoal series. If there are $n$ subgoals, the size of abstract states is $n+1$. The agent is in a first abstract state~$z_0$ before a subgoal is achieved. Then, the abstract state $z_0$ transits to $z_1$ when the subgoal $sg_0$ is achieved. This means the aggregation of episodes until subgoal $sg_0$ transits into $z_0$. The aggregated episodes change dynamically every trial because of the policy with randomness and learning. As the learning progresses, the aggregated episodes become fixed. The value over abstract states is distributed to the values of states of the trajectory. Note that the trajectories for updating the values are different from those of distributed values. The updated value function is not used for the current trial but for the next trials. An image of dynamic trajectory aggregation is shown in Fig.~\ref{fig:concept}.

In the figure, a circle is a state, and the aggregated states are in each gray background area. There are two abstract states in the case of a single subgoal. The bold circles express the states with which the designer deals. The number of bold circles in Fig.~\ref{fig:dta} is much lower than Fig.~\ref{fig:sarsa-rs-concept}. ``S" and ``G" in the circles are a start and a goal, respectively. Fig.~\ref{fig:dta} shows that the episode is separated into two sub-episodes, and each of them corresponds to the abstract states. 

\subsubsection{Accumulated Reward Function}
We clearly define the reward transformation function~$r_h$ because our method only updates the achievements of subgoals as abstract states. A set of abstract states is part of the semi-Markov decision process~(SMDP)\cite{Sutton:1999:MSF:319103.319108}. The transition between an abstract state and another consists of multiple actions. The value function in SMDP is written as
\begin{eqnarray*}
	V_{g}(z) = E \left\{ \sum_{i=0}^{k-1} \gamma^ir_{t+1+i} + \gamma^k V_{g} (z') | \varepsilon(g, z, t) \right\}
\end{eqnarray*}
, where $k$ is the duration of the abstract state $z$, and $\varepsilon$ is the event of the aggregation function $g$ being initiated in state $z$ at time $t$. Therefore, we describe this formally as $r_h = \sum_{i=0}^{k-1} \gamma^{i} r_i$, where $k$ is the duration until subgoal achievement. The function accumulates rewards with discount~$\gamma$. Depending on the policy at the time, $k$ is varied dynamically. This follows n-step temporal difference~(TD) learning~\cite{Sutton1998} because there are transitions between an abstract state $z_i$ and another one $z_{i+1}$. 
Algorithm~\ref{alg:subgoal_online_learning} shows the whole process of SARSA-RS with subgoal-based dynamic trajectory aggregation. $\alpha_v$ and $\gamma_v$ are hyper-parameters, that is, the learning rate and discount factor for updating the value function over abstract states.

\begin{algorithm}
	\caption{SARSA-RS with subgoal-based dynamic trajectory aggregation} 
	\label{alg:subgoal_online_learning}
	\KwData{$t=0, V(z; \theta), sg_i \in \{SG, \prec \}$}
	Initialize $\theta$ \\
	$z \leftarrow z_0, i \leftarrow 0, r_h \leftarrow 0$ \\
	Select $a$ by $\pi$ at $s$ \\
	\Repeat{terminal condition}{
		Take $a$ and observe $s'$ and $r$\\
		$z' \leftarrow filter(s')$\\
		$t \leftarrow t+1 $ \\
		\If{$equal(s', sg_{i+1})$ or $equal(s', g)$}{
			\If{$equal(s', g)$}{
				$\delta = r_h + \gamma_v^t r - V(z; \theta)$ \\
			}\Else{
				$\delta = r_h + \gamma_v^tV(z'; \theta) - V(z; \theta)$ \\
			}
			$\theta \leftarrow \theta + \alpha_v \delta_\theta \Delta V(z; \theta)$ \\
			$t \leftarrow 0, i \leftarrow i + 1, r_h \leftarrow 0$ \\
		}
		$r_h \leftarrow r_h + \gamma^tr$ \\
		$F(z,z') = \gamma V(z') - V(z)$ \\
		Select $a'$ by $\pi$ \\
		Update value function with $r + F(z,z')$ \\
		$s \leftarrow s'; a \leftarrow a', z \leftarrow z'$
	}
\end{algorithm}
In Algorithm~\ref{alg:subgoal_online_learning}, the method is involved between lines 6-16.  The value function over abstract states is parameterized by $\theta$. If $s'$ equals $sg_{i+1}$ or a goal state~$g$, $\theta$ is updated by an approximate multi-step TD method \cite{Sutton1998}. If $s'$ is $g$, the target is $r$, not $V(z';\theta)$ because $g$ is a terminal state.

\section{Experiment}
In this section, we show that our method improves learning efficiency in a simulation.
The task was a social navigation task~\cite{7989037}. There is a single robot and a single human moving on the basis of an ORCA policy~\cite{10.1007/978-3-642-19457-3_1}, and they aim for their own individual goals. On the way to the goal, the robot must avoid collision with the human. The task succeeds when the robot reaches the goal. In practice, the robot travels around for over 25 steps, or collides with the human, and then the task fails. The start and goal position of the robot and human, shown in Fig.~\ref{fig:cn}, are fixed through learning, respectively. 

We compared CADRL shaped by our dynamic trajectory based aggregation with CADRL~(DTA)and CADRL shaped by naive reward shaping(NRS).
NRS is based on potential-based reward shaping. The potential function $\Phi(s)$ outputs a scalar value $\eta$ just when an agent has visited a subgoal state. The potential function is written formally as follows.
\begin{eqnarray}
	\Phi(s) = \left \{\begin{array}{ll}
		\eta&s = sg \\
		0&s\neq sg
	\end{array}
	\right.
\end{eqnarray}
Informally, NRS shapes the rewards of $\eta$ only generated for subgoals with potential-based reward shaping. The two differences from our method are that NRS has a fixed potential, and the positive potential is only for the subgoals.  
We evaluated the success rate, collision rate, navigation time, and total reward as performance indicators~\cite{Heiden2020Social}. 
The navigation time is the number of steps for the robot to reach the goal. The total reward is the sum of the rewards that the robot acquires. The results are averaged in 10 learning for the performance during learning, and in 500 test runs for one after learning. The success and collision rate is the ratio of success and collision in 500 test runs which the parameters are not updated, respectively. 
The policy is $\epsilon$ greedy, and $\epsilon$ decayed linearly from 0.5 to 0.4 in the all episodes. For the value network, the learning rate was 0.001, and the discount factor was 0.9. The network architecture was the same as that of \cite{7989037}. To initialize the value network, the ORCA policy generated 300 episodes. The parameter $eta$ was 1, which was the same as the reward value at the goal. All the experiments were conducted with a PC [Ryzen 9 5950X, 16 cores(3.4GHz), 128 GB of memory].
\subsection{Experimental Results}
\begin{figure}[tb]
	\centering
	\begin{subfigure}[b]{0.23\textwidth}
		\centering
		\includegraphics[width=\textwidth, bb=0.000000 0.000000 1074.000000 1073.000000]{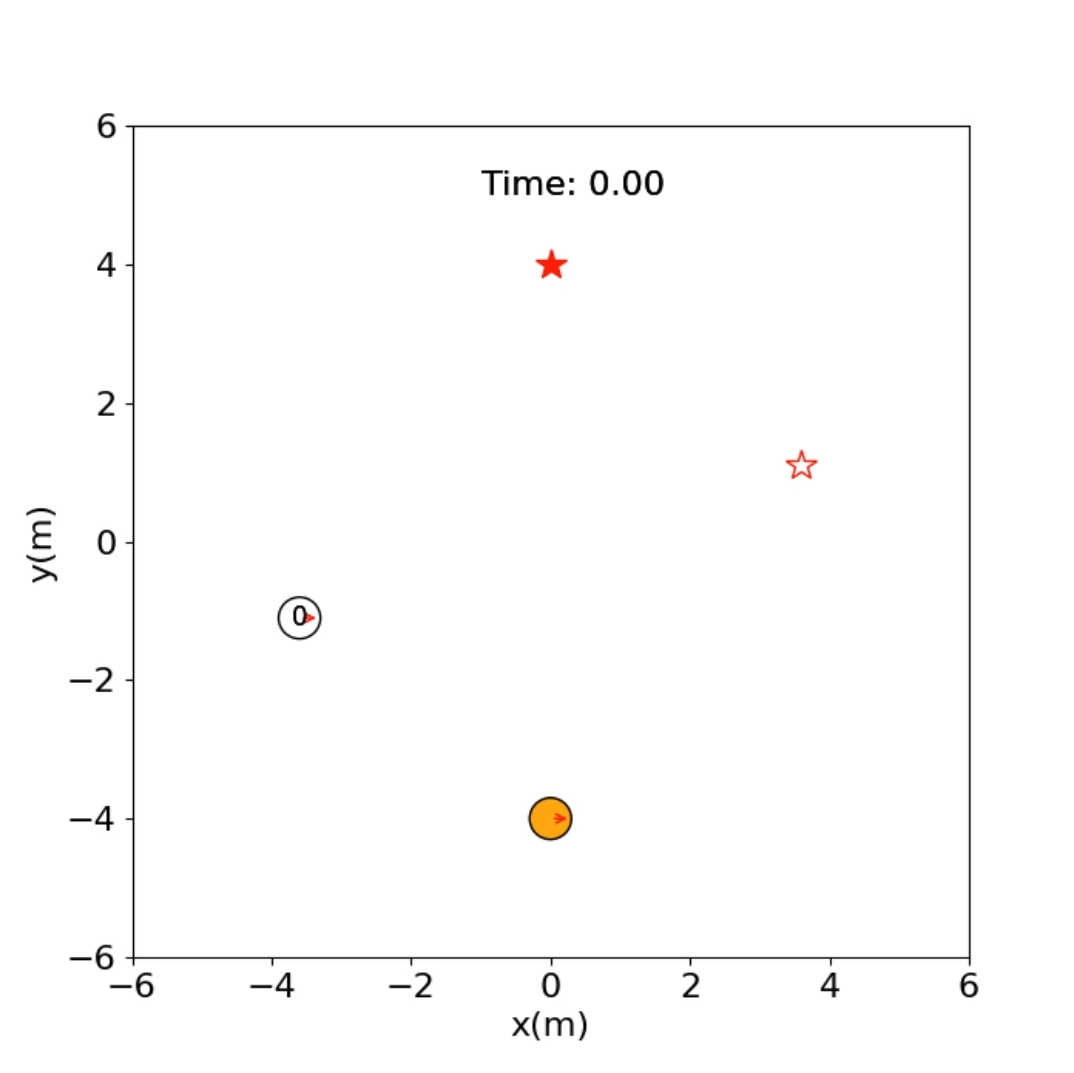}
		\caption{Position pattern}
		\label{fig:cn}
	\end{subfigure}
	\begin{subfigure}[b]{0.23\textwidth}
		\centering
		\includegraphics[width=\textwidth, bb=0.000000 0.000000 289.945129 290.425170]{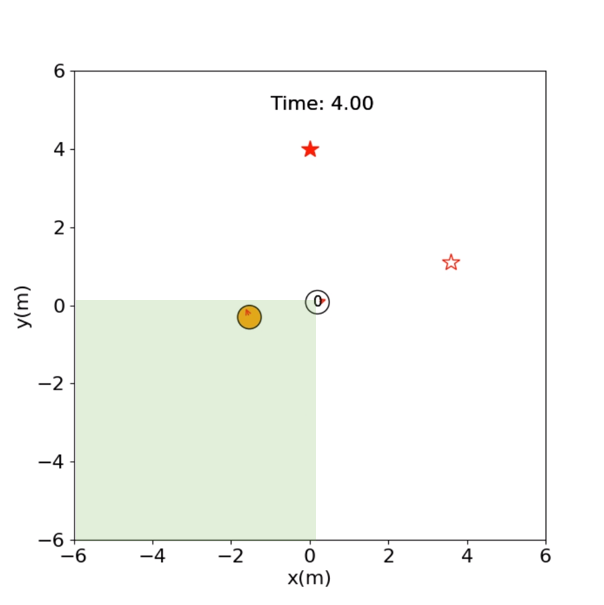}
		\caption{Subgoal}
		\label{fig:sn_subgoal}
	\end{subfigure}
	\caption{Crowd navigation. Yellow circle is robot, white circle is human. Red and white star are goals of robot and human, respectively.}
\end{figure}

We used the subgoal as shown in Fig. \ref{fig:sn_subgoal}. The subgoal as the green shaded area means that the robot should pass behind the human, which was the condition to be satisfied. We came up with this subgoal inspired by~\cite{Scandolo2011anthropomorphic, Middlemist1976Personal}. We used the relative position vector from the human to the robot and the velocity vector of the human. The angle between both vectors was obtained by using an arc tangent function. We set the angles to 135 degrees at minimum and of 225 degrees at maximum. When the robot and the human satisfied the condition, the subgoal was achieved. DTA and NRS used this subgoal.

\begin{figure}
	\centering
		\begin{subfigure}[b]{0.5\textwidth}
		\centering
		\includegraphics[width=\textwidth]{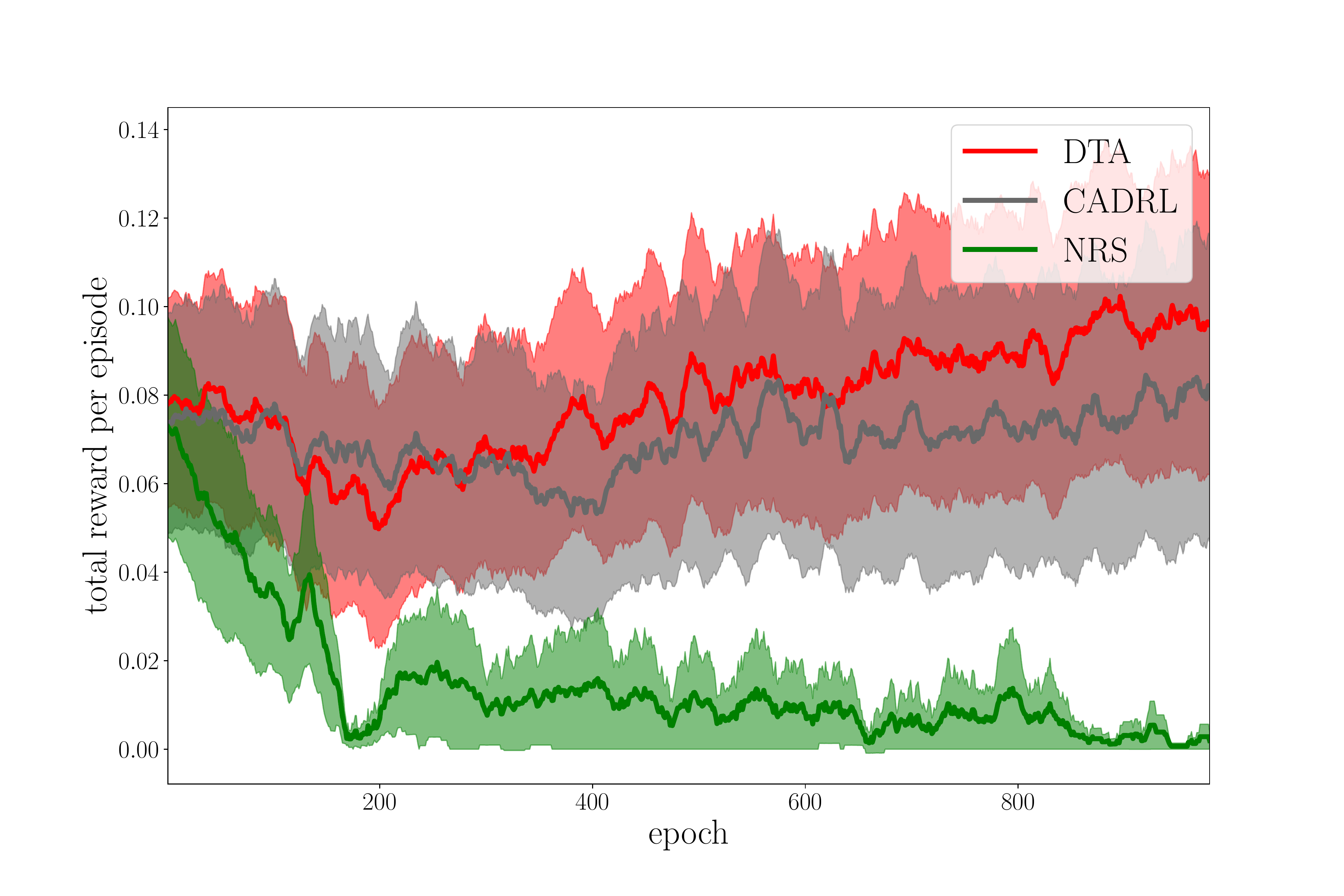}
		\caption{Total reward}
		\label{fig:total_reward_lc}
	\end{subfigure}
	\hfill
	\begin{subfigure}[b]{0.5\textwidth}
		\centering
		\includegraphics[width=\textwidth]{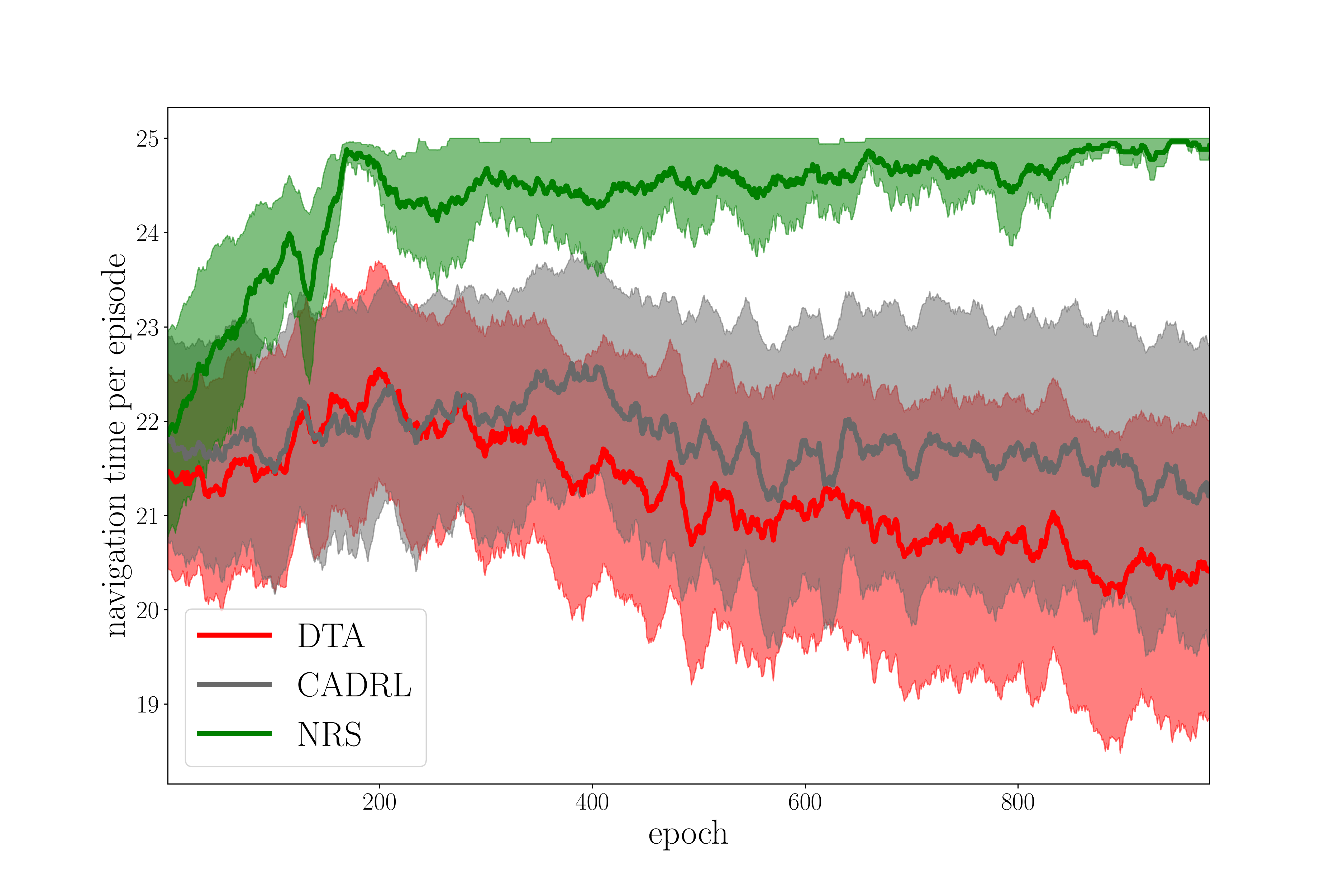}
		\caption{Navigation time}
		\label{fig:nav_time_lc}
	\end{subfigure}
	\caption{Learning curves. Line and shaded areas are mean and standard errors for 5 learnings, respectively.}
	\label{fig:learning_curves}
\end{figure}

Fig.~\ref{fig:learning_curves} shows the learning curves for the navigation time and total reward. 
As shown in Fig.~\ref{fig:nav_time_lc}, DTA could learn the navigation with higher total reward and with less steps than CADRL through the almost whole learning as shown in Fig.~\ref{fig:total_reward_lc}. DTA stably reached 0.08 about 100 episodes earlier than CADRL in the total reward. The collision rate of both methods was always zero. NRS had the worst performance through the whole learning. This shows the difficulty of designing of potentials with subgoals. Our method makes subgoals easy to be provided for accelerating learning.
Table~\ref{tab:pal} shows the performance after learning. The performance of DTA was the best among the three methods in success rate, navigation time, and total reward. The collision rates of all methods are zero.
\begin{table}
	\centering
	\begin{tabular}{| c | c | c | c | c |}
		\hline
		method & suc. rate & nav. time & col. rate & total reward\\ 
		\hline
		DTA & *0.5 & *17.2 & 0.00 & *0.19\\  
		CADRL& 0.4 & 18.4 & 0.00 & 0.17\\
		NRS& 0.1 & 22.7 & 0.00 & 0.04\\
		\hline
	\end{tabular}
	\caption{\label{tab:pal} Performance after learning. Suc. rate is success rate, nav. time is navigation time, col. rate is collision rate. We averaged 500 episodes generated by learnt policy. * indicates the best performance.}
\end{table}

\subsection{Discussion}
Fig.~\ref{fig:learning_curves} shows the large standard errors. CADRL did not always learn well. When the learning failed, the success rate and total reward were zero for almost every episode. In the success, the success rate was one, and the total reward was about 0.3. The navigation time was 25 steps as the max number of steps, and about 10 steps as the success. The successes and failures are mixed in Fig.~\ref{fig:learning_curves}, so the standard errors became large. DTA was affected by the base RL algorithm because it depends on the state-value function learned by acquired rewards. If the base RL algorithm did not succeed, DTA always outputs zero rewards, so learning could not be improved. This is also why DTA did not improve learning efficiency of the base RL algorithm in the begging of learning. This is our future work.\par
We have not made clear how many subgoals and what subgoals a lot of humans give most easily. Moreover, we have not validated what the easiest way for humans to give knowledge is yet. There are many types of human knowledge transfer, including trajectories, preferences, policies, advice, interactive feedback, and subgoals. Our method has high flexibility to several of these types. For example, in the case of trajectories, we can extract subgoals from trajectories. This is useful because it makes trajectories of failure useful information for teaching an agent. In the case of interactive feedback, we can also extract subgoals. We can use the states with positive feedback as subgoals. We consider the classification of knowledge transfer for appropriate tasks to be meaningful. To this end, we will conduct a large participant experiment. \par
In the current experiment, we used a single subgoal for which an agent passes behind a human. This biased the agent to adhere to social compliance. DTA does not always tell an agent to adhere to any social compliance because it is based on PBRS. PBRS transforms a reward function with the original optimal policy remaining. DTA is not appropriate when a reward function is required to change for adherence to social compliance. DTA can bias learning for acquiring a trajectory with social compliance if there are trajectories generated by an optimal policy.

\section{CONCLUSION}
Though reinforcement learning which selects an action at low computational cost, is useful for social navigation, it takes an enormous number of iterations until acquiring a behavior policy. This may become a barrier to social implementation. We proposed a reward shaping method with subgoals that accelerates learning. To simplify a teacher's input, SARSA-RS, which simultaneously updates the potentials through learning, is used as the basis. We incorporated subgoal knowledge into reward shaping by dynamic trajectory-based aggregation. We evaluated the method in a social navigation task in which a robot and a human were present. We compared our method with CADRL and a naive reward shaping. From the results, our method improved the learning efficiency in comparison with both CADRL and naive reward shaping.
\addtolength{\textheight}{-12cm}   

\bibliographystyle{IEEEtran.bst}
\bibliography{Bibliography.bib}

\end{document}